\documentclass[10pt,journal,compsoc]{IEEEtran}
\ifCLASSOPTIONcompsoc
  \usepackage[nocompress]{cite}
\else
  \usepackage{cite}
\fi
\ifCLASSINFOpdf
\else
\fi
\usepackage{url}

\usepackage{amsmath,amsfonts}
\usepackage{subfigure}
\usepackage{nicefrac}       
\usepackage{subfigure}
\usepackage{booktabs}
\usepackage{bm}
\usepackage{multirow}
\usepackage{orcidlink}
\usepackage{xcolor} 

\definecolor{ccr}{RGB}{0,0,255}  
\usepackage{hyperref}
\hypersetup{hypertex=true,
		colorlinks=true,
		linkcolor=ccr,
		urlcolor=black,
		citecolor=ccr}

\begin{document}
%
\title{Exploiting Inter-Image Similarity Prior for Low-Bitrate Remote Sensing Image Compression}
%
%
%
%

\author{Junhui Li${^{\orcidlink{0000-0001-9143-1321}}}$ and Xingsong Hou${^{\orcidlink{0000-0002-6082-0815}}}$
\thanks{Manuscript received xx, 2024; This work was supported by the National Natural Science Foundation of China under Grant 62272376.  \textit{(Corresponding authors: Xingsong Hou.)}	The authors are with the School of Information and Communications Engineering, Xi’an Jiaotong University, Xi’an 710049, China (e-mail: mlkkljh@stu.xjtu.edu.cn; houxs@mail.xjtu.edu.cn).	
}}

%
%

\markboth{Journal of \LaTeX\ Class Files,~Vol.~14, No.~8, August~2015}%
{Shell \MakeLowercase{\textit{et al.}}: Bare Demo of IEEEtran.cls for Computer Society Journals}
%




\IEEEtitleabstractindextext{%
\begin{abstract}
Deep learning-based methods have garnered significant attention in remote sensing (RS) image compression due to their superior performance. Most of these methods focus on enhancing the coding capability of the compression network and improving entropy model prediction accuracy. However, they typically compress and decompress each image independently, ignoring the significant inter-image similarity prior. In this paper, we propose a codebook-based RS image compression (Code-RSIC) method with a generated discrete codebook, which is deployed at the decoding end of a compression algorithm to provide inter-image similarity prior. Specifically, we first pretrain a high-quality discrete codebook using the competitive generation model VQGAN. We then introduce a Transformer-based prediction model to align the latent features of the decoded images from an existing compression algorithm with the frozen high-quality codebook. Finally, we develop a hierarchical prior integration network (HPIN), which mainly consists of Transformer blocks and multi-head cross-attention modules (MCMs) that can query hierarchical prior from the codebook, thus enhancing the ability of the proposed method to decode texture-rich RS images. Extensive experimental results demonstrate that the proposed Code-RSIC significantly outperforms state-of-the-art traditional and learning-based image compression algorithms in terms of perception quality. The code will be available at \url{https://github.com/mlkk518/Code-RSIC/}.
\end{abstract}

}

\maketitle

\IEEEdisplaynontitleabstractindextext

\IEEEpeerreviewmaketitle
\IEEEraisesectionheading{\section{Introduction}}\label{sec:intro}
\IEEEPARstart{R}{emote} sensing (RS) imagery is widely used in various fields, including environmental monitoring \cite{7887698}, urban planning \cite{rs15153895}, and disaster management \cite{lorenzo2018earth}. The extensive use of RS imagery results in vast volumes of data, necessitating efficient compression techniques. Current compression methods have significant challenges in keeping pace with the rapidly increasing data volumes, prompting a pressing need for advanced compression algorithms capable of handling such large datasets effectively. \par

Traditional image compression methods, such as JPEG2000 \cite{taubman2002jpeg2000}, BPG \cite{bpg2017}, WebP \cite{bib_webP}, and versatile video coding (VVC) \cite{VVC2021}, have been crucial for the efficient storage and transmission of image data. Despite their widespread use, these standards have significant limitations. Firstly, the block-based hybrid coding techniques they employ require sequential encoding and decoding processes, which can result in undesirable blocking or ringing artifacts in the decoded images. Secondly, the complex interdependencies among their hand-crafted modules pose challenges for the joint optimization of the entire coding algorithm. Additionally, with the rise of diverse applications with specific needs in RS images, exploring more advanced compression methods has become urgent.

With the rise of deep learning, numerous learning-based image compression algorithms have emerged \cite{47602_Johannes, cheng2020learned, he2022elic, guo2021causal, li2020learning, 10378746}. For example, Ballé \textit{et al.} \cite{47602_Johannes} developed a variational image compression method using a hyperprior entropy model. Following this, Cheng \textit{et al.} \cite{cheng2020learned} introduced a discretized Gaussian mixture model to better capture the prior distribution of latent features, thereby enhancing the accuracy of the entropy model prediction. Then, He \textit{et al.} \cite{he2022elic} proposed an innovative uneven channel-conditional adaptive grouping scheme to effectively leverage prior information in latent representations for more accurate entropy prediction.
In practice, RS image compression presents significant challenges in achieving high rate-distortion (RD) performance due to its complex contextual and spectral information. Consequently, some researchers have focused on developing learning-based RS image compression algorithms \cite{zhang2023global, pan2023coupled, xiang2023remote, fu2023remote}. For instance, Zhang \textit{et al.} \cite{zhang2023global} enhanced network feature extraction capabilities by incorporating a multi-scale attention module, introducing global priors, and using anchored-stripe attention to refine the entropy model. Pan \textit{et al.} \cite{pan2023coupled} employed generative adversarial networks (GANs) to decode image content and complex textures separately, achieving low-bitrate RS image compression. Wang \textit{et al.} \cite{pan2023coupled} proposed to leverage historical ground station images as references for on-orbit compression, reducing redundancy in RS images and enhancing compression efficiency. However, it is still hard to achieve impressive perception quality, especially at low bitrates. \par
More recently, attention has turned towards text-guided image compression algorithms to improve the perceptual quality of decoded images \cite{lee2024neural, jiang2023multi, osti_10490383}. For example, Lee \textit{et al.} \cite{lee2024neural} utilized text-adaptive encoding and trained with joint image-text loss to develop a compression framework, which presented impressive performance. Eric \textit{et al.} \cite{osti_10490383} proposed to utilize text descriptions with side information to generate high-fidelity decoded images that preserve both semantics and spatial structure. However, the performance of these algorithms heavily depends on the accuracy of the guiding text, which significantly limits their practical application.

While these algorithms have showcased competitive RD performance in image compression, few have explored leveraging inter-image similarity prior to enhance compression performance. 
For the past few years, codebook-based learning methods have garnered significant attention due to their powerful ability to provide additional prior knowledge in a wide range of image tasks \cite{10533792, qin2023blind, jia2024generative, bilal2021fast}. In image compression tasks, Mao \textit{et al.} \cite{10533792}  proposed to represent an input image as vector quantization (VQ) indices by determining the nearest codeword, followed by applying lossless compression techniques to compress these indices into bitstreams. Jia \textit{et al.} \cite{jia2024generative} employed VQVAE  \cite{van2017neural} to compress the image into the latent space, which was then compressed by a transform coding module with a code-prediction-based latent supervision for subsequent bitstream coding. \par

\begin{figure}[!htbp]
	\centering
	\includegraphics[width=0.45\textwidth]{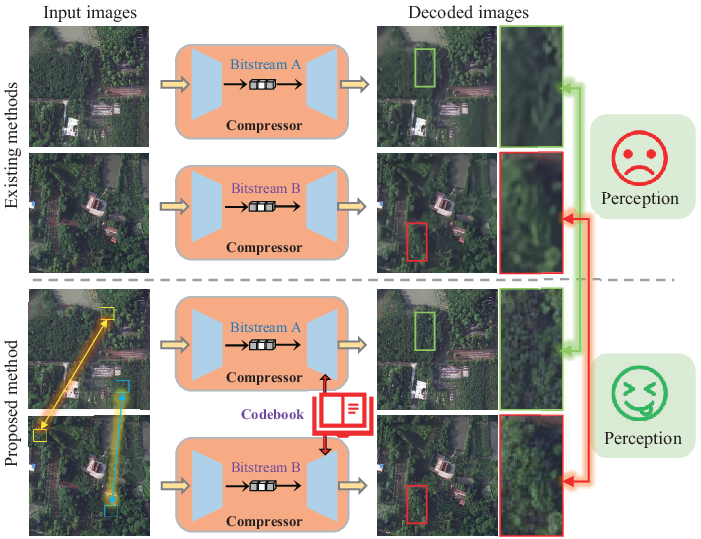}
	\caption{Illustration of the difference between existing image compression algorithms and the proposed method. Here the role of the codebook is to provide inter-image similarity prior to the decoding end of the compressor.}
	\label{Fig: piple_com}
\end{figure}
Despite achieving impressive compression performance, these methods generally require retraining the entire model when applied to new target datasets. Furthermore, the priors provided by these methods offer insufficient guidance for decoding texture-rich RS images. As shown in Fig. \ref{Fig: piple_com}, existing compression methods typically compress and decompress each image independently, ignoring the inter-image similarity prior \cite{RTAVAKOLI201710, He_2023_CVPR}, such as the image regions highlighted by the yellow and blue boxes in the figure. In contrast, we aim to generate a high-quality codebook that codes these similarities at the decoding end of an image compression algorithm in advance. This approach not only eliminates the need to transmit additional bitstreams but also produces perception-friendly decoded images.

In this paper, we propose a novel codebook-based RS image compression (Code-RSIC) method, which leverages a pre-trained high-quality codebook to enhance the decoding performance of an existing image compression algorithm. Specifically,  we first train a high-quality discrete codebook using a large number of high-quality RS images with the competitive generation model VQGAN. Next, we introduce a Transformer-based prediction module to bridge the gap between the latent features of the initial decoded images and the derived codebook. Finally, we utilize Transformer blocks and multi-head cross-attention modules (MCMs) to develop a hierarchical prior integration network (HPIN), which enables hierarchical prior information queries from the codebook, resulting in improved decoded RS images.

The primary contributions of this paper can be summarized as follows:
\begin{itemize}
	\item[$\bullet$] We propose the Code-RSIC method for RS image compression, which leverages prior information from a pre-trained high-quality codebook. This method can be easily integrated as a common component to enhance the decoded image quality of existing image compression algorithms.
	\par
	\item[$\bullet$]  To the best of our knowledge, we are the first to utilize the inter-image similarity prior to enhance the decoding performance in RS image compression, offering a new perspective on decoding texture-rich RS images. \par
	\item[$\bullet$] By employing the competitive Transformer blocks and MCMs, we develop HPIN to better utilize the codebook prior for decoding perception-friendly RS images.  
	\par
	\item[$\bullet$] Extensive experimental results demonstrate that the proposed Code-RSIC significantly outperforms state-of-the-art traditional and learning-based image compression algorithms in terms of perceptual quality, especially at extremely low bitrates.
	\par
\end{itemize}

The remainder of this paper is organized as follows. In Section \ref{Sec: realated}, we review related works on learning-based image compression and codebook learning. In Section \ref{Sec:Method}, we introduce the proposed method in detail, and the experimental results are reported in Section \ref{Sec: Experiments}. Finally, we conclude this paper in Section \ref{Sec:Conclusion}.

\section{Related Work} \label{Sec: realated}
\subsection{Learning-Based Image Compression}
With the rapid development of deep learning technologies, learning-based image compression has garnered wide attention in recent years \cite{zhang2023global, qian2020learning, he2022elic, pan2023coupled}. In \cite{toderici2015variable}, the authors explored applying LSTMs to image compression and achieved significant improvements. Subsequently, many studies have focused on developing learning-based image compression techniques. For instance, Ballé \textit{et al.} \cite{47602_Johannes} introduced a side information prior for entropy model prediction and proposed to incorporate a zero-mean Gaussian distribution model to capture the spatial dependencies of the latent representation. Following this, numerous researchers \cite{cheng2020learned, qian2022entroformer, liu2023learned, 10379598, 10516594} aimed to develop competitive encoding and decoding networks or more accurate entropy models for improved RD performance.

For instance, Cheng \textit{\textit{et al.}} \cite{cheng2020learned} resorted to a discretized Gaussian mixture model (GMM) to parameterize the distributions of latent representations, developing a more accurate entropy model. Given the Transformer's strong capability to capture global dependencies, Qian \textit{\textit{et al.}} \cite{qian2022entroformer} integrated the Transformer to enhance the prediction accuracy of the entropy model. Additionally, Zou \textit{et al.} \cite{zou2022devil} devised a symmetrical Transformer (STF) framework composed solely of Transformer blocks for both the down-sampling encoder and up-sampling decoder development. Furthermore, He \textit{\textit{et al.}} \cite{he2022elic} developed an uneven channel-conditional adaptive coding scheme after observing the energy compaction characteristic in learned-based image compression, and proposed an innovative efficient learned image compression (ELIC) method.
Recently, Liu \textit{\textit{et al.}} \cite{liu2023learned} took advantage of the local feature-capturing capabilities of convolutional neural networks (CNNs) and the long-range dependencies enabled by Transformers. They devised an efficient Transformer-CNN Mixture (TCM) block, which was integrated into their encoder and decoder network designs. They also developed a channel-wise entropy model to achieve greater accuracy in entropy modeling.
Furthermore, in \cite{10379598}, the authors transformed the input image into the frequency domain and developed two-path bit streams to separately compress the low and high frequencies. In \cite{10516594}, Transformer blocks and patch-based local attention were adopted to enhance the representation ability of the encoding network and entropy prediction accuracy, respectively. 

In contrast to these algorithms, we aim to deploy the codebook at the decoding end of an image compression algorithm, using it as prior information for improved RS image decoding. This approach allows our method to be easily integrated as a component of existing image compression algorithms without requiring retraining of the compression algorithm.

\subsection{Codebook Learning}
The VQVAE~\cite{van2017neural} pioneered the concept of a highly compressed codebook through a vector-quantized autoencoder model. Building on this foundation, VQGAN~\cite{esser2021taming} innovatively integrates adversarial and perceptual losses to further improve perceptual quality, achieving impressive performance. Afterward, several codebook-based learning methods have been reported, demonstrating their great power in computer version tasks, such as face restoration \cite{zhou2022towards, gu2022vqfr}, image dehazing \cite{wu2023ridcp}, and image compression \cite{10533792}. For instance, Zhou \textit{et al.} \cite{zhou2022towards}  leveraged the competitive power of a VQGAN-based discrete codebook and developed a Transformer-based prediction network for blind face restoration. Similarly, Wu \textit{et al.} \cite{wu2023ridcp} first pretrained a VQGAN on a large-scale high-quality dataset to derive a discrete codebook, which was then utilized as a robust prior for network development in real image dehazing. 

Inspired by recent advancements in codebook-based learning algorithms, our goal is to offer a generic component for existing compression algorithms using discrete codebooks, thereby facilitating their implementation in low-bitrate RS image compression. In contrast to recent VQGAN-based approaches~\cite{zhou2022towards, wu2023ridcp}, we integrate the discrete codebook into our method using a multi-head cross-attention mechanism and Transformer-based hierarchical feature fusion. This integration allows our method to fully leverage the benefits of a pre-trained codebook without requiring additional bits.

\section{Proposed Method} \label{Sec:Method}
In this section, we will begin with an overview of Code-RSIC. Subsequently, we will provide detailed explanations of Stage I, which involves learning the codebook; Stage II, which focuses on looking up the codebook; and Stage III, which involves fusing the features of the learned codebook. \par
\begin{figure}[!htbp]
	\centering
	\subfigure [\fontsize{7}{10}\selectfont Matching features of images A and B]
	{	
		\begin{minipage}[t]{0.45\textwidth}
			\centering
			\includegraphics[scale=0.85]{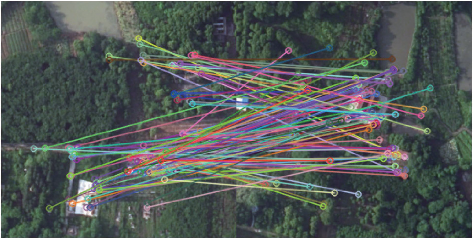}
		\end{minipage}	
	}%
	
	\subfigure [\fontsize{7}{10}\selectfont Clustering results for similar features of images A and B]
	{
		\begin{minipage}[t]{0.45\textwidth}
			\centering
			\includegraphics[scale=0.45]{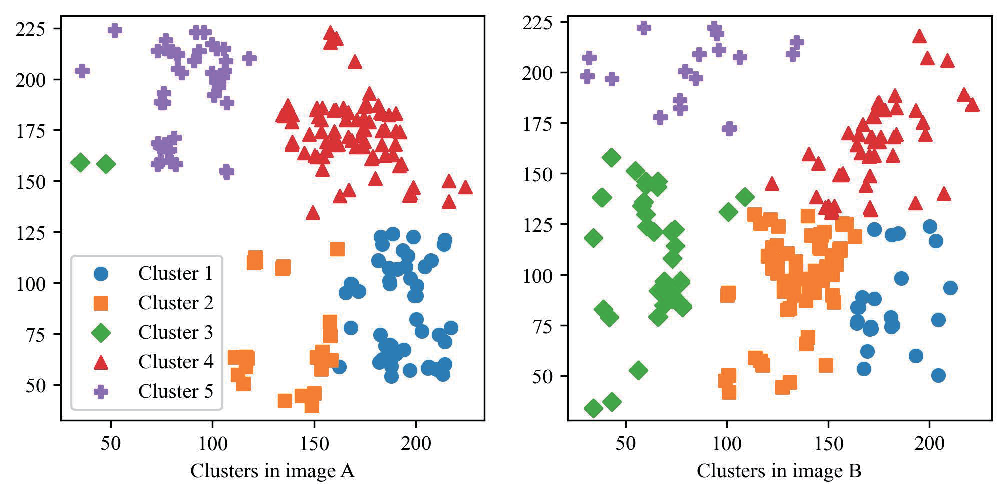}
		\end{minipage}	
	}%
	\caption{Visualization of inter-image similarity prior between two images.}
	\label{Fig: Simlary_images}
\end{figure}
Generally, in image compression, leveraging image similarity is pivotal for efficient bitstream saving. Existing compression methods \cite{2021ITIP_liu, kim2022joint, qian2022entroformer, 9521684, 10516594} generally exploit local and non-local similarities within individual images. Fig. \ref{Fig: Simlary_images} illustrates the similarities between the two images. Specifically, Fig. \ref{Fig: Simlary_images}(a) depicts feature matching results obtained using ORB \cite{rublee2011orb} and the brute force matching algorithm \cite{bradski2000opencv}. Fig. \ref{Fig: Simlary_images}(b) presents clustering results for similarities using the K-Means algorithm \cite{pedregosa2011scikit} on the matched features. These results highlight the existence of inter-image similarity prior across different images,  which has also been confirmed in the studies \cite{RTAVAKOLI201710, He_2023_CVPR}. Motivated by this observation, we propose incorporating a pre-trained codebook at the decoding end of an image compression algorithm. It is expected to utilize the inter-image similarity prior in the codebook to encourage one image compression algorithm to decode texture-rich RS images. 

\begin{figure*}[t]
	\centering
	\includegraphics[width=0.98\textwidth]{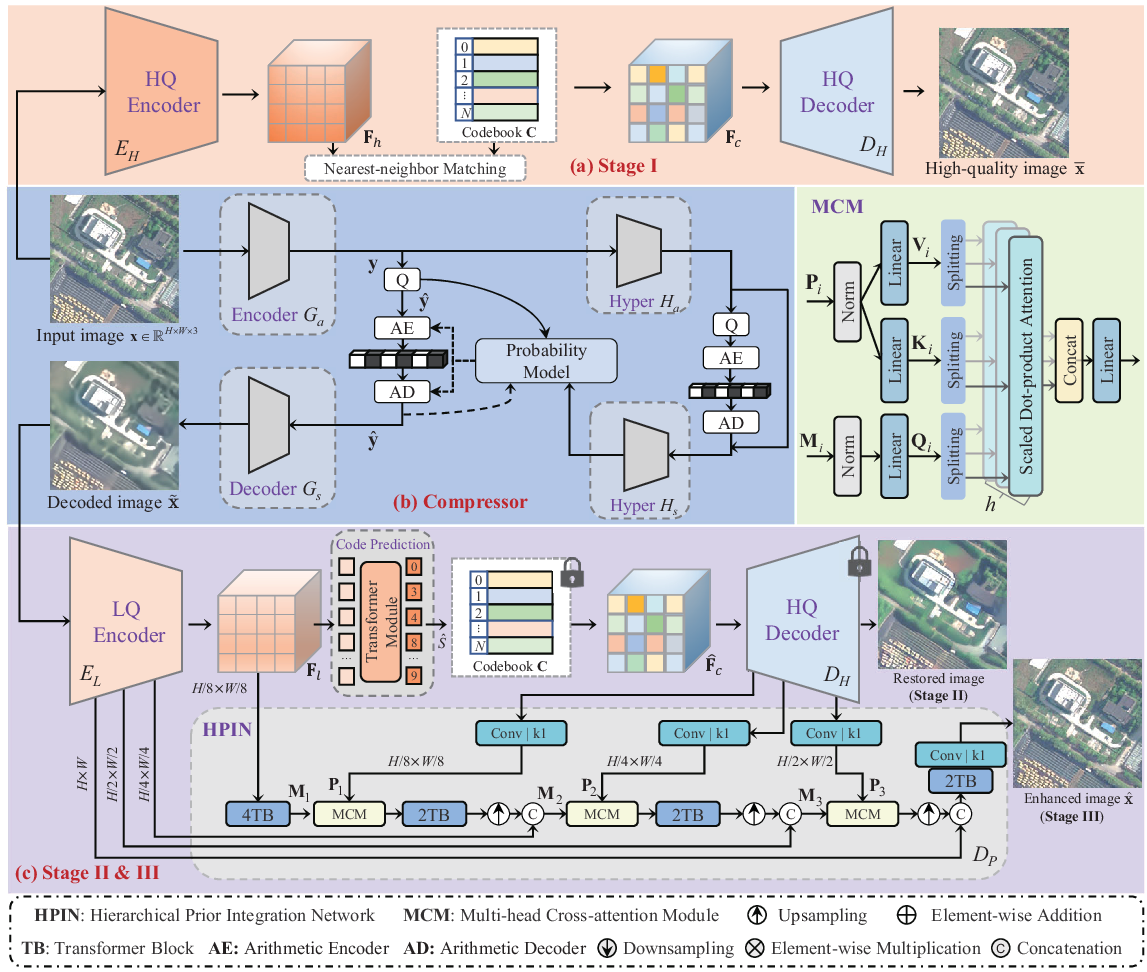}
	\caption{Framework of the proposed Code-RSIC. We first learn a discrete codebook $\mathbf{C}$ and the decoder $D_H$ to store high-quality visual parts of RS images via self-reconstruction learning. Then with frozen codebook and decoder $ D_H $, we introduce a Transformer module \cite{zhou2022towards} for code sequence prediction, modeling the global RS image composition of decoded image from the compressor ELIC \cite{he2022elic}. Besides, HPIN is developed to bridge the information flow from LQ encoder $E_L$ to decoder $D_H$ and query prior information from the frozen $D_H$.}
	\label{Fig:framework}
\end{figure*}
\subsection{Overview of Code-RSIC}
In our development, as shown in Fig. \ref{Fig:framework}(b), we adopt ELIC \cite{he2022elic} as the compressor. Specifically, for an input image \(\mathbf{x} \in \mathbb{R}^{W \times H \times 3}\), the encoder \(G_a\) extracts the latent representation \(\mathbf{y}\), followed by a quantization operation \(\operatorname{Q}(\cdot)\) to derive \(\mathbf{\hat{y}}\). Then \(\mathbf{\hat{y}}\) is encoded by a hyper entropy coder into a bitstream. On the decoding side, we can obtain the decoded image \({\tilde{\mathbf{x}}}\) by feeding \(\mathbf{\hat{y}}\) into the decoder \(G_s\). The process can be described as
\begin{equation}
	\begin{aligned}
		\mathbf{y} &= G_a(\mathbf{x}; \Omega), \\
		\mathbf{\hat{y}} &= \operatorname{Q}(\mathbf{y}), \\
		{\tilde{\mathbf{x}}} &= G_s(\mathbf{\hat{y}}; \Theta),
	\end{aligned}
\end{equation}
where \(G_a\) and \(G_s\) refer to the neural networks with \(\Omega\) and \(\Theta\) representing their respective network parameters. To facilitate the arithmetic coding of \(\hat{\mathbf{y}}\), hyper networks \(H_a\) and \(H_s\) are typically designed to derive the probability model.

Additionally, to achieve rate control, the compression networks are jointly optimized for bitrate (R) and reconstruction mean squared error  \(\operatorname{MSE}(\cdot)\). The loss function can be expressed as
\begin{align}
	\mathcal{L}_\text{RD} = \operatorname{R} + \lambda \operatorname{MSE}(\mathbf{x}, G_s(\mathbf{\hat{y}})), \label{Eq: loss_com}
\end{align}
where $\lambda$ is a hyper-parameter to control the trade-off between the bitrate and distortion.\par

As illustrated above, ELIC aims to consume fewer bitrates to obtain decoded images with higher quality. Considering the inter-image similarity prior among various RS images, we aim to develop a high-quality codebook \(\mathbf{C}\) to provide prior information that assists ELIC in decoding texture-rich RS images without additional bitstreams.
As shown in Fig.~\ref{Fig:framework}(a), we first build upon VQ~\cite{van2017neural} and VQGAN technology~\cite{esser2021taming} to create a discrete codebook and its corresponding decoder \(D_H\) with high-quality RS images. Subsequently, as depicted in Fig. \ref{Fig:framework}(c), we introduce a Transformer module~\cite{zhou2022towards} for accurate prediction of code combinations from the decoded RS images.
Furthermore, we develop a hierarchical prior integration network (HPIN) with Transformer blocks \cite{vaswani2017attention} and multi-head cross-attention modules (MCMs) to better utilize the hierarchical prior information from the codebook to obtain improved RS images. Accordingly, the training of our method is structured into three distinct stages.

\subsection{Stage I: Codebook Prior Learning}
\label{sec:codebook_learning}
\subsubsection{Codebook Development}
To reduce the uncertainty of the low-quality decoded images to the high-quality image reconstruction and obtain texture-rich RS images, we first pretrain a quantized autoencoder to learn a high-quality codebook.
As shown in Fig.~\ref{Fig:framework}(a), the encoder \(E_H\) and decoder \(D_H\) are adopted as in \cite{zhou2022towards}. The input image \(\mathbf{x} \in \mathbb{R}^{H \times W \times 3}\) is first embedded as a latent feature \(\mathbf{F}_h \in \mathbb{R}^{u \times v \times d}\) by the encoder \(E_H\). Then, as in VQGAN-based approaches~\cite{esser2021taming, wu2023ridcp}, the nearest-neighbor matching operation is employed within the learnable codebook \(\mathbf{C} = \{\mathbf{c}_n \in \mathbb{R}^d\}_{n=0}^{N-1}\) to generate the quantized feature \(\mathbf{F}_c \in \mathbb{R}^{u \times v \times d}\) and the corresponding code sequence \(S \in \{0, 1, \ldots, N-1\}\). This process can be described as
\begin{equation}
	\begin{aligned}
		\mathbf{F}_c^{(u,v)} &= \mathop\text{arg min}\limits_{\mathbf{c}_n\in \mathbf{C}} \|\mathbf{F}_h^{(u,v)} - \mathbf{c}_n\|, \\
		\quad
		S^{(u,v)} &= \mathop\text{arg min}\limits_{n} \|\mathbf{F}_h^{(u,v)} - \mathbf{c}_n\|.
	\end{aligned}
	\label{eq:nn}
\end{equation}
The decoder \(D_H\) then reconstructs the high-quality RS image \(\bar{\mathbf{x}}\) using \(\mathbf{F}_c\).

The $ N $ sequence \(S\) forms a new latent discrete representation that specifies the respective code index of the learned codebook, \textit{i.e.}, \(\mathbf{F}_c^{(u,v)}=\mathbf{c}_n\) when \(S^{(u,v)}=n\). 

\subsubsection{Loss Functions}
To train the codebook, we use three loss functions: reconstruction loss \(\mathcal{L}_{rec}\), perceptual loss \(\mathcal{L}_{per}\), and adversarial loss \(\mathcal{L}_{adv}\), which can be expressed as follows:
\begin{equation}
	\begin{aligned}
		\mathcal{L}_{rec} &= \|\mathbf{x}-\bar{\mathbf{x}}\|_1, \quad \\
		\mathcal{L}_{per} &= \|\operatorname{Vgg}(\mathbf{x})-\operatorname{Vgg} (\bar{\mathbf{x}})\|^2, \quad \\
		\mathcal{L}_{adv} &= \log \operatorname{D}(\mathbf{x}) + \log(1-\operatorname{D}(\bar{\mathbf{x}})),
	\end{aligned}
	\label{eq:image_level_loss}
\end{equation}
where $\rm Vgg(\cdot)$ refers to the feature extractor of VGG19~\cite{simonyan2015very}, and $\operatorname{D}(\cdot)$ denotes the discriminator network as adopted in \cite{zhou2022towards}.

Additionally, as image-level loss is not sufficiently constrained in updating the codebook items, we employ an intermediate code-level loss \(\mathcal{L}_{cl}\), as adopted in \cite{esser2021taming, zhou2022towards}, to reduce the distance between the codebook \(\mathbf{C}\) and the input feature \(\mathbf{F}_h\). This process can be written as
\begin{equation}
	\mathcal{L}_{cl} = \|\operatorname{SG}[\mathbf{F}_h] - \mathbf{F}_c\|^2 + \alpha\|\mathbf{F}_h - \operatorname{SG}[\mathbf{F}_c]\|^2,
	\label{eq:code_feat_loss}
\end{equation}
where \(\operatorname{SG}[\cdot]\) denotes the stop-gradient operator, and the weighting factor \(\alpha = 0.25\) is used to balance the update rates of the encoder and the codebook. Additionally, as the quantization operation in Eq.~\eqref{eq:nn} is non-differentiable, the straight-through gradient estimator \cite{courbariaux2015binaryconnect, he2022elic} is employed to enable gradient propagation from the decoder to the encoder. Therefore, the total loss function for codebook prior learning can be derived by
\begin{equation}
	\mathcal{L}_{s1} = \mathcal{L}_{rec}+\mathcal{L}_{per}+\mathcal{L}_{cl}+\lambda_{1}\mathcal{L}_{adv},
	\label{eq:stage1_loss}
\end{equation}
where $\lambda_{1}$ is an adaptive weight. Following the study \cite{esser2021taming}, here it can be obtained by
\begin{equation}
	\lambda_{1} =  \frac { \Vert \nabla _ { D _ { H } } [ \mathcal{L} _ { r e c } ] \Vert }  { \Vert \nabla _ { D _ { H } } [ \mathcal{L}_{adv} ]  \Vert+ \epsilon }, \label{Eq: Stage1_loss_lambda}
\end{equation}
where \(\nabla D_H [\cdot]\) represents the gradient of its input concerning the final layer of the decoder \(D_H\), and \(\epsilon = 10^{-4}\) is utilized to ensure training stability.

\subsection{Stage II: Transformer-Based Codebook Lookup} \label{sec:transformer}
\subsubsection{Codebook Lookup}
Due to texture loss in the decoded RS images of ELIC, the nearest-neighbor matching in Eq.~\eqref{eq:nn} often fails to find accurate codes for RS image restoration (see Fig. \ref{Fig: vis_VQGAN}). To address this problem, as shown in Fig. \ref{Fig:framework}(c), we introduce a Transformer module \cite{zhou2022towards} behind the encoder \(E_L\) to capture global interdependencies, enhancing code prediction. The network parameters of the codebook \(\mathbf{C}\) and decoder \(D_H\) are frozen. The encoder \(E_L\), which has the same network structure as \(E_H\), loads the trained network parameters of \(E_H\) from Stage I and is then fine-tuned for RS image restoration.

Concretely, the decoded image $\tilde{\mathbf{x}}$ of ELIC is first passed through the encoder \(E_L\) to obtain the feature \(\mathbf{F}_l \in \mathbb{R}^{u \times v \times d}\). Subsequently, the Transformer module \cite{zhou2022towards} is adopted to predict the \(N\) code sequence \(\hat{S} \in \{0, 1, \cdots, N-1\}\) as probabilities in terms of \(N\) code items. The corresponding \(N\) code items can then be derived from the learned codebook according to the predicted code sequence, resulting in the quantized feature \(\hat{\mathbf{F}}_c \in \mathbb{R}^{u \times v \times d}\). This, in turn, can generate the restored RS image through the fixed decoder \(D_H\).
\subsubsection{Loss Functions}
The Transformer module and encoder $E_L$ are respectively trained and finetuned for RS image restoration, while the model parameters of codebook $\mathbf{C}$ and decoder $D_H$ are frozen. Two loss functions are employed to achieve model training. Specifically, cross-entropy loss \(\mathcal{L}_{ce}\) is introduced to predict the code token. L2 loss \(\mathcal{L}_{qf}\) is developed to minimize the gap between the decoded RS image feature \(\mathbf{F}_l\) and the quantized feature \(\mathbf{F}_c\) from the codebook, which is beneficial for the learning of token prediction. The two functions can be formulated as
\begin{equation}
	\begin{aligned}
		\mathcal{L}_{ce} &= \sum\limits_{n=0}^{N-1} -S_n\log(\hat{S}_n), \\
		\mathcal{L}_{qf} &= \|\mathbf{F}_l - \text{SG}(\mathbf{F}_c)\|^2,
	\end{aligned}
	\label{eq:code_seq_loss}
\end{equation}
where the ground truth (GT) of the latent code \(S\) is obtained from the pre-trained autoencoder in Stage I. The quantized feature \(\mathbf{F}_c\) is then retrieved from the codebook according to \(S\). The total loss function to achieve codebook lookup can be written as follows:
\begin{equation}
	\mathcal{L}_{s2} = \mathcal{L}_{qf} + \lambda_{2}\mathcal{L}_{ce},
	\label{eq:stage2_loss}
\end{equation}
where $\lambda_{2}$ is set to 0.5, as adopted in \cite{zhou2022towards}, in the proposed Code-RSIC.

\subsection{Stage III: Hierarchical Prior Integration}
\label{sec:cpt}
\subsubsection{Prior Integration}
In this section, as shown in Fig. \ref{Fig:framework}(c), we propose the HPIN, composed primarily of Transformer blocks and MCMs, to effectively integrate the learned prior and the intermediate features of the encoder \(E_L\). HPIN facilitates the fusion of features from \(E_L\) and the high-quality prior of the decoder \(D_H\), enabling the low-quality decoded images to obtain more texture details. 
Considering that a single-scale prior feature may not adequately adapt to various decoded images, we start by halving the channels for each scale of prior feature dimensions using a \(1 \times 1\) convolution layer to reduce complexity. Therefore, we can obtain hierarchical prior features \( \mathbf{P} = \{ \mathbf{P}_1, \mathbf{P}_2, \mathbf{P}_3 \} \) from \(D_H\).
In each MCM, we use cross-attention mechanism \cite{NEURIPS2019_01894d6f} to compute interactions between the obtained features \(\mathbf{P}\) and the intermediate features \( \mathbf{M} = \{ \mathbf{M}_1, \mathbf{M}_2, \mathbf{M}_3 \} \) of HPIN to facilitate feature fusion. \par

Specifically, as shown in MCM of Fig.~\ref{Fig:framework}, given the intermediate $i$-th feature $\mathbf{M}_i \in \mathbb{R}^{\hat{H} \times \hat{W} \times \hat{C}}$, where $i \in \{1, 2, 3\}$, we reshape it into tokens $\vec {\mathbf{M}}_i \in \mathbb{R}^{\hat{H}\hat{W} \times \hat{C}}$, where $\hat{H} \times \hat{W}$ is the spatial resolution, and $\hat{C}$ denotes the channel dimension. After a normalization layer, we then linearly project $\vec {\mathbf{M}}_i$ into $\mathbf{Q}_i \in \mathbb{R}^{\hat{H}\hat{W} \times \tilde{C}}$ (Query). Similarly, we reshape the prior feature $\mathbf{P}_i \in \mathbb{R}^{\hat{H} \times \hat{W} \times C'}$ into $\vec {\mathbf{P}}_i \in \mathbb{R}^{\hat{H} \hat{W} \times C'}$. Through a normalization layer, two linear layers project $\vec {\mathbf{P}}_i$ into $\mathbf{K}_i \in \mathbb{R}^{\hat{H} \hat{W} \times \tilde{C}}$ (Key) and $\mathbf{V}_i \in \mathbb{R}^{\hat{H} \hat{W} \times \tilde{C}}$ (Value). Thus, the cross-attention result can be formulated as follows:
\begin{equation}
	\begin{gathered}
		\mathbf{Q}_i = \vec {\mathbf{M}}_i \mathbf{W}_i^Q ,~~ \mathbf{K}_i =  \vec {\mathbf{P}}_i \mathbf{W}_i^K, ~~ \mathbf{V}_i =  \vec{\mathbf{P}}_i \mathbf{W}_i^V,\\
		{\operatorname{Attention}}(\mathbf{Q}_i, \mathbf{K}_i, \mathbf{V}_i) = {\operatorname{Softmax}} \left(\frac{\mathbf{Q}_i \mathbf{K}_i^{\rm {T}}}{\sqrt{{\tau}_i}}\right) \mathbf{V}_i,
		\label{eq:HIM}
	\end{gathered}
\end{equation}
where \(\mathbf{W}_i^Q \in \mathbb{R}^{\hat{C} \times \tilde{C}}\), \(\mathbf{W}_i^K \in \mathbb{R}^{{C}' \times \tilde{C}}\), and \(\mathbf{W}_i^V \in \mathbb{R}^{{C}' \times \tilde{C}} \) denote the learnable parameters of the linear layers without bias, and \({\tau}_i\) represents a scaling factor. \par
Additionally, to capture relevant information across different subspaces, the multi-head attention mechanism \cite{vaswani2017attention, 10041164} is adopted to split channels into $h$ scaled dot-product attention, which can be defined as follows:
\begin{equation}
	\begin{gathered}
		\mathbf{H}_j = \operatorname{Attention}(\mathbf{Q}_i^j, \mathbf{K}_i^j, \mathbf{V}_i^j), \\
		\operatorname{Multihead}(\mathbf{Q}_i, \mathbf{K}_i, \mathbf{V}_i) = \operatorname{Concat}(\mathbf{H}_1, \ldots, \mathbf{H}_h)\mathbf{W}^O,
	\end{gathered}
\end{equation}
where $\mathbf{Q}^j_i, \mathbf{K}^j_i, \mathbf{V}^j_i \in \mathbb{R}^{\hat{H} \hat{W} \times \frac{\tilde{C}}{h}}$ denote the splited $j$-th subspaces, $j \in \{1, 2, \ldots, h\}$ and $\mathbf{W}^O \in \mathbb{R}^{\tilde{C} \times \tilde{C}}$ refers to a linear mapping. 


\subsubsection{Loss Functions}
To optimize the model, the codebook $\mathbf{C}$ and decoder $ D_H $ are frozen. In addition to the code-level loss \(\mathcal{L}_{s2}\) from Stage II, we also utilize the image reconstruction loss \(\mathcal{L}^{'}_{rec}\), perception loss \(\mathcal{L}^{'}_{per}\), and adversarial loss \(\mathcal{L}^{'}_{adv}\), which can be formulated as follows:
\begin{equation}
	\begin{aligned}
		\mathcal{L}^{'}_{rec} &= \|\mathbf{x}-\hat{\mathbf{x}}\|_1, \quad \\
		\mathcal{L}^{'}_{per} &= \|\rm Vgg(\mathbf{x})-\rm Vgg (\hat{\mathbf{x}})\|^2, \quad \\
		\mathcal{L}^{'}_{adv} &= \log \rm {D}(\mathbf{x}) + \log(1- \rm {D}(\hat{\mathbf{x}})),
	\end{aligned}
	\label{eq:image_level_loss_S3}
\end{equation}
where $\hat{\mathbf{x}}$ refers to the final enhanced RS image. Thereafter, the total loss can be obtained by weighting the above loss terms, formulated as
\begin{equation}
	\mathcal{L}_{s3} = \mathcal{L}_{s2} + \mathcal{L}^{'}_{rec}+\mathcal{L}^{'}_{per}+  {\lambda}_{3}\mathcal{L}^{'}_{adv},
	\label{eq:stage3_loss}
\end{equation}
where  $ {\lambda}_{3}$ is calculated by
\begin{equation}
	\lambda_{3} =  \frac { \Vert \nabla _ { D _ { P } } [ \mathcal{L}^{'} _ { r e c } ] \Vert }  { \Vert \nabla _ { D _ { P } } [ \mathcal{L}^{'}_{adv} ]  \Vert+ \epsilon }, \label{Eq: Stage3_loss_lambda}
\end{equation}
where \(\nabla D_P [\cdot]\) refers to the gradient of its input with respect to the final layer of the HPIN.

\section{Experimental Results} \label{Sec: Experiments}
\subsection{Experimental Settings}
\subsubsection{Datasets}
To evaluate the performance of the proposed Code-RSIC, we conduct experiments on the rural images of the RS image dataset LoveDA \cite{10516594}. The LoveDA contains 5,987 high-quality RS images with 166,768 annotated objects from Nanjing, Changzhou, and Wuhan cities. There are 2,713 images for urban scenes and 3,274 images for rural scenes. each with a resolution of $1024\times1024 \times 3$ in ``PNG" format. \par
\begin{figure}[!htbp]
	\centering
	\includegraphics[width=0.45\textwidth]{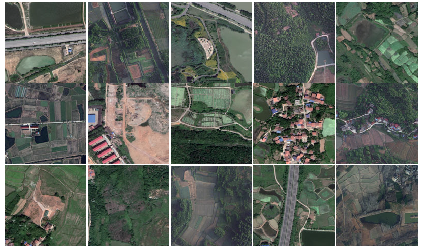}
	\caption{Some cropped images from the training set of the LoveDA dataset.}
	\label{Fig: training_examples}
\end{figure}

\begin{table}[!htbp]
	\centering
	\caption{Dataset partitioning for rural scenes of LoveDA}
	\begin{tabular}{cccc}
		\toprule
		Datasets & Cities & Regions & \#Images \\
		\hline
		\multirow{4}{*}{Training} & \multirow{3}{*}{Nanjing} & Pukou & 320 \\
		&     & Gaochun & 336 \\
		&     & Lishui & 336 \\
		& Wuhan & Jiangxia & 374 \\
		\cmidrule(lr){2-4}	
		\multirow{3}{*}{Testing} & Nanjing & Jiangning & 336 \\
		& \multirow{2}{*}{Changzhou} & Liyang & 320 \\
		&     & Xinbei & 320 \\
		\bottomrule
	\end{tabular}%
	\label{tab: dataset_div}%
\end{table}%
Tab. \ref{tab: dataset_div} summarized the dataset partitioning of rural scenes in LoveDA, and some examples of the training set are provided in Fig. \ref{Fig: training_examples}. The rural images have rich textures, which is a challenge for existing image compression algorithms to obtain a high RD performance. In this experiment, we randomly crop 27,319 image patches with the resolution of $256\times 256 \times 3$ from its training set for model training and randomly crop 300 images for its testing set with the same resolution for model performance evaluation. It should be noted that, as adopted in \cite{10516594}, several augmentation operations, such as random horizontal, vertical flip, and random crop are employed to enrich the diversity of the training dataset.
\begin{figure*}[!htbp]
	\centering
	\subfigure 
	{
		\centering
		\begin{minipage}[t]{0.24\textwidth}
			\includegraphics[scale=0.6]{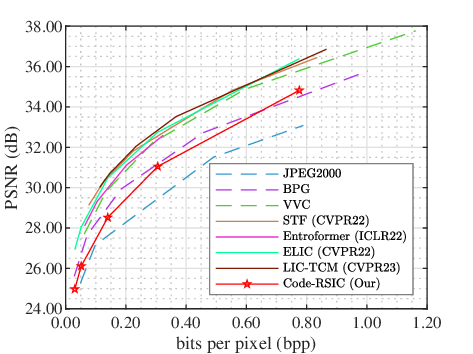}
		\end{minipage}	
	}%
	\subfigure 
	{
		\centering
		\begin{minipage}[t]{0.24\textwidth}
			\includegraphics[scale=0.6]{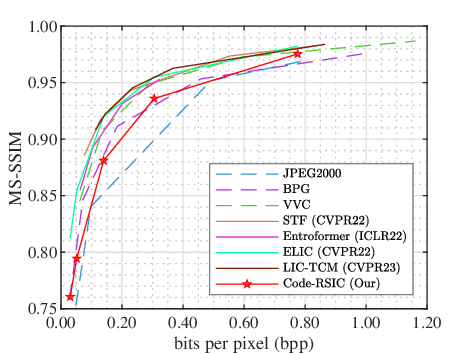}
		\end{minipage}	
	}%
	\subfigure 
	{
		\centering
		\begin{minipage}[t]{0.24\textwidth}
			\includegraphics[scale=0.6]{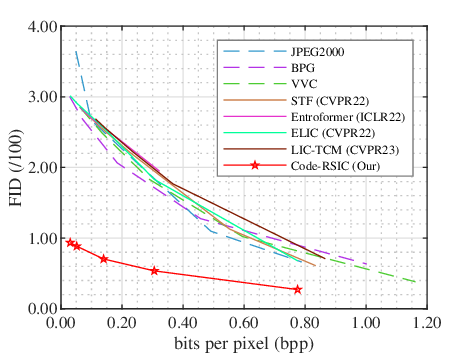}
		\end{minipage}	
	}%
	\subfigure 
	{
		\centering
		\begin{minipage}[t]{0.24\textwidth}
			\includegraphics[scale=0.6]{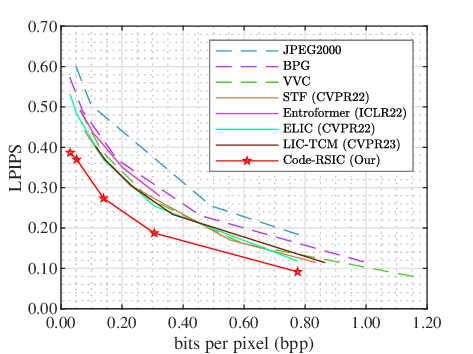}
		\end{minipage}	
	}%
	\centering
	\caption{Rate-distortion and rate-perception curves of several image compression algorithms on the testing set of LoveDA.}
	\label{Fig: RD_LoveDA}
\end{figure*}
\subsubsection{Implementation Details}
In our experiments, we employ ELIC \cite{he2022elic} to compress images of the training set for facilitating model training. Specifically, we utilize five pre-trained ELIC models with parameters $\lambda \in \{4, 8, 32, 100, 450\} \times 10^{-4}$,  as in Eq. (\ref{Eq: loss_com}), to decode the images separately, which serve as the initial decoded images for training our proposed model. For the codebook items, inspired by \cite{zhou2022towards}, the number \(N\) and the code dimension \(d\) of the codebook are set to 1024 and 512, respectively. In the $i$-th MCM, the number of heads $h_i$ is set to $\{4,2,2\}$ for $i=1,2,3$.

For training optimization, we utilize the Adam optimizer \cite{kingma2014adam} with parameters $\beta_1 = 0.9$ and $\beta_2 = 0.999$. In Stage I, the initial learning rate is set to 1e-5, and the total number of iterations is 1600K. After 30K iterations, the learning rate is scheduled using cosine annealing \cite{loshchilov2017sgdr}, with a minimum value of 1e-6.
In Stage II, the total epochs are set to 500K, starting with an initial learning rate of 1e-5. The learning rate is halved when reaching 400K and 450K epochs. In Stage III, the learning rate remains fixed at 1e-5. Stages I and II have a batch size of 8 and Stage III has a batch size of 2. Early stopping is further employed to prevent overfitting and expedite training.

To assess the effectiveness of our proposed Code-RSIC, we measure coding bitrates in bits per pixel (bpp) and use two perception metrics:  Fréchet inception distance (FID) \cite{heusel2017gans} and learned perceptual image patch similarity (LPIPS) \cite{zhang2018unreasonable} for evaluating perceptual quality. The lower of these two metrics indicates the better perceptual quality of the corresponding image. Additionally, we employ two quantitative metrics: peak signal-to-noise ratio (PSNR) and multi-scale structural similarity (MS-SSIM) \cite{wang2003multiscale} to assess image distortion.

\subsection{Quantitative Comparison}
To verify the performance of the proposed Code-RSIC,  we compare it with four traditional image compression standards: JPEG2000\footnote{\href{https://kakadusoftware.com/}{https://kakadusoftware.com}} \cite{taubman2002jpeg2000}, BPG\footnote{\href{https://bellard.org/bpg/}{https://bellard.org/bpg}} \cite{bpg2017}, WebP \cite{bib_webP}, and VVC (YUV 444)\footnote{\href{https://vcgit.hhi.fraunhofer.de/jvet/VVCSoftware_VTM/-/releases/VTM-16.0}{https://vcgit.hhi.fraunhofer.de/jvet/VVCSoftware\_VTM}} \cite{VVC2021}. Additionally, we compare it with competitive learning-based image compression algorithms such as 
Entrorformer\footnote{\href{https://github.com/damo-cv/entroformer}{https://github.com/damo-cv/entroformer}} \cite{qian2022entroformer}, STF\footnote{\href{https://github.com/Googolxx/STF}{https://github.com/Googolxx/STF}} \cite{zou2022devil},  ELIC\footnote{\href{https://github.com/VincentChandelier/ELiC-ReImplemetation}{https://github.com/VincentChandelier/ELiC-ReImplemetation}} \cite{he2022elic}, and LIC-TCM\footnote{\href{https://github.com/jmliu206/LIC\_TCM?tab=readme-ov-file}{https://github.com/jmliu206/LIC\_TCM?tab=readme-ov-file}} \cite{liu2023learned}. 

Fig. \ref{Fig: RD_LoveDA} illustrates the rate-distortion and rate-perception curves of the proposed Code-RSIC and comparison algorithms on the testing set of LoveDA. The results show that Entrorformer and ELIC achieve competitive perception results in terms of FID and LPIPS metrics among these comparison algorithms, respectively. However, it is evident that the proposed Code-RSIC outperforms all compression algorithms significantly in the two metrics. This highlights Code-RSIC's superior capability in achieving high-quality perceptual decoding of RS images. 

In addition, it is observed that the proposed Code-RSIC presents comparable decoding performance to BPG in terms of PSNR and MS-SSIM, indicating that Code-RSIC shows mediocre performance in quantitative metrics. In summary, the results demonstrate that while the proposed method has limited performance in objective quality improvement, it excels in enhancing perceptual quality.

\begin{figure*}[!htbp]
	\centering
	\subfigure [\fontsize{7}{10}\selectfont Decoding results on image ``4286"]
	{
		\centering
		\begin{minipage}[t]{1\textwidth}
			\includegraphics[scale=1.2]{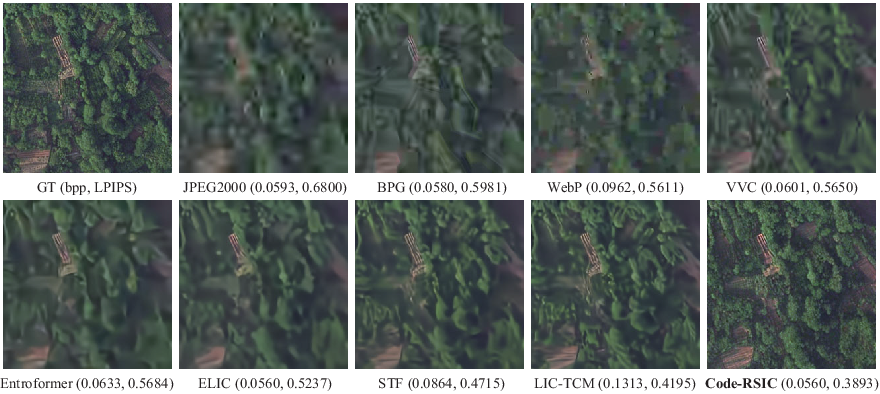}
		\end{minipage}	
	}%
	\vspace{-1.5mm}
	\subfigure [\fontsize{7}{10}\selectfont  Decoding results on image ``4483"]
	{
		\centering
		\begin{minipage}[t]{1\textwidth}
			\includegraphics[scale=1.2]{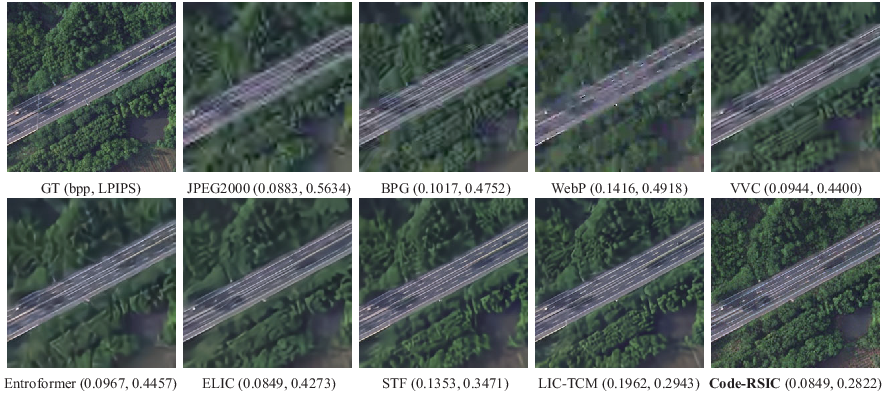}
		\end{minipage}	
	}%
	\centering
	\caption{Visual performance of several image compression algorithms on two images of the testing set of LoveDA at low bitrates ($\lambda=0.0008$).}
	\label{Fig: vis_low_bpp}
\end{figure*}

\subsection{Qualitative Comparison}
\begin{figure*}[!htbp]
	\centering
	\subfigure[\fontsize{7}{10}\selectfont  Decoding results on image ``4216"]{
		\centering
		\begin{minipage}[t]{0.98\textwidth}
			\includegraphics[scale=1.2]{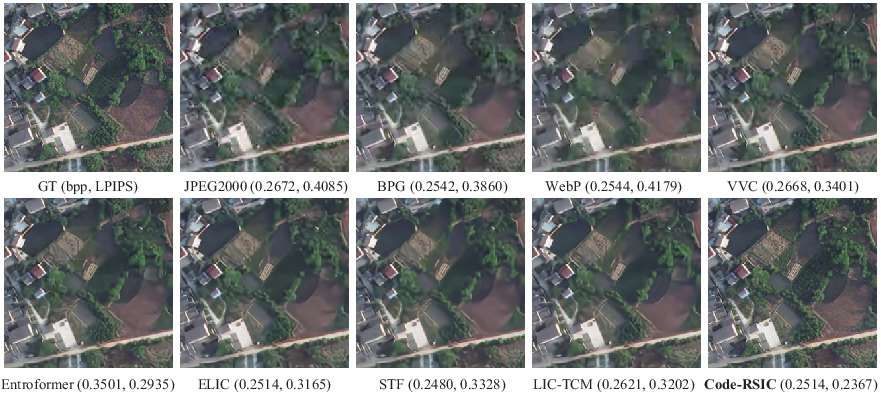}
		\end{minipage}	
	}%
	\vspace{-1.5mm}
	\subfigure[\fontsize{7}{10}\selectfont  Decoding results on image ``4455"]{
		\centering
		\begin{minipage}[t]{0.98\textwidth}
			\includegraphics[scale=1.2]{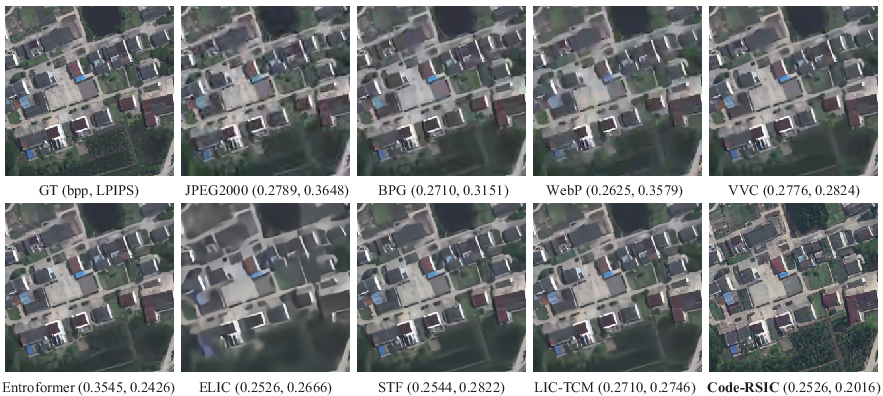}
		\end{minipage}	
	}%
	\centering
	\caption{Visual performance of several image compression algorithms on two images of the testing set of LoveDA at high bitrates ($\lambda=0.0032$).}
	\label{Fig: vis_high_bpp}
\end{figure*}

To visualize the performance of various image compression algorithms, Figs. \ref{Fig: vis_low_bpp} and \ref{Fig: vis_high_bpp} present decoded images from the testing set of LoveDA at low and high bitrates, respectively. At low bitrates, as illustrated by image ``4286" in Fig. \ref{Fig: vis_low_bpp}(a), traditional algorithms like JPEG2000, BPG, and WebP exhibit noticeable artifacts and blurring effects, severely limiting image quality. Even the state-of-the-art VVC struggles to decode texture-rich images effectively.
Turning to learning-based compression algorithms, all suffer from varying degrees of blurring and loss of texture information in decoded images. For instance, although the bpp value of LIC-TCM is 2.34 times higher than Code-RSIC, as seen in image ``4483" of Fig. \ref{Fig: vis_low_bpp}(b), it fails to produce satisfactory results. This comparison underscores the competitive performance of Code-RSIC in decoding texture-rich RS images.

At high bitrates, as depicted in Fig. \ref{Fig: vis_high_bpp}(a), while VVC shows impressive results compared to traditional algorithms, it still falls short in decoding perception-friendly images when compared to Code-RSIC. Despite consuming higher bitrates, LIC-TCM suffers from severe blurring effects. Moreover, Fig. \ref{Fig: vis_high_bpp}(b) further validates the effectiveness of Code-RSIC in decoding texture-rich RS images. In summary, the decoded images at low and high bitrates demonstrate that Code-RSIC consistently delivers competitive performance in decoding texture-rich RS images, even at extremely low bitrates, making it well-suited for RS image compression tasks.

\subsection{Ablation Studies}
In this section, we first explore the impact of the three stages in the proposed Code-RSIC. We then evaluate the decoding performance of Code-RSIC with and without the codebook. 

\subsubsection{Impact of Stages}
\begin{figure*}[!htbp]
	\centering
	\includegraphics[width=0.96\textwidth]{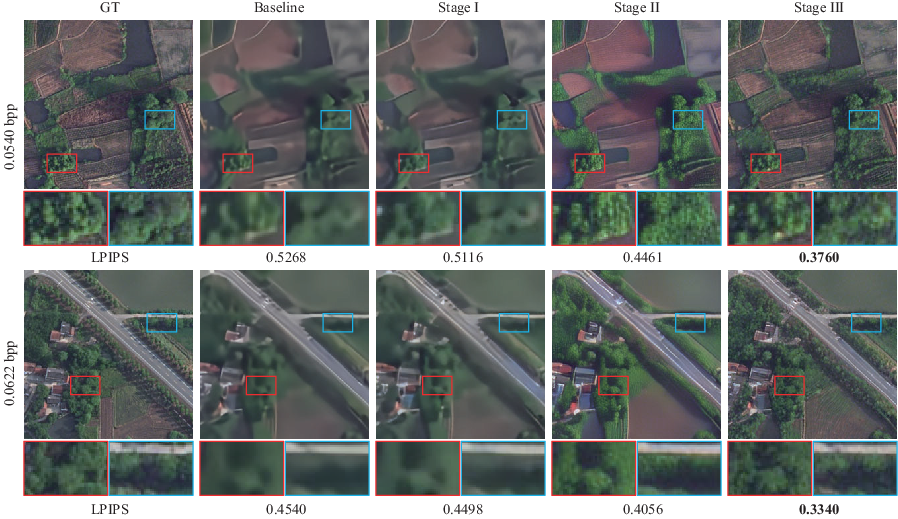}
	\caption{Decoded images of the Stages I, II, and III under $\lambda=0.0008$. LPIPS is adopted to indicate the perception quality of the decoded images.}
	\label{Fig: vis_VQGAN} 
\end{figure*}

To visualize the decoded results of the three stages, Fig. \ref{Fig: vis_VQGAN} presents the decoded images of the baseline (\textit{i.e.}, ELIC) and the decoded images of each stage. From the results, it is clear that Stage I has no obvious positive effect on decoding high-quality RS images. This is because, in Stage I, we only adopt high-quality RS images for generating the codebook, and there are significant gaps between the GT images and the decoded image of the baseline, which highly limits its final reconstruction performance. After introducing a prediction strategy to adjust the gap between the high-quality GT images and the decoded images in Stage II, benefiting from the nearest-neighbor matching operation, one can observe that the decoded images present more texture-rich details than those in Stage I. However, it is evident that the matching difference results in noticeable color loss. \par
To address this issue, in Stage III, with the help of the proposed HPIN, the features of the decoded images and the predicted codebook from Stage II are fused for final image reconstruction. The results show that the decoded images present impressive perceptual quality compared to the baseline, which strongly indicates the effectiveness of the developed HPIN.
\begin{figure}[!htbp]
	\centering
	\subfigure 
	{
		\centering
		\begin{minipage}[t]{0.24\textwidth}
			\includegraphics[scale=0.6]{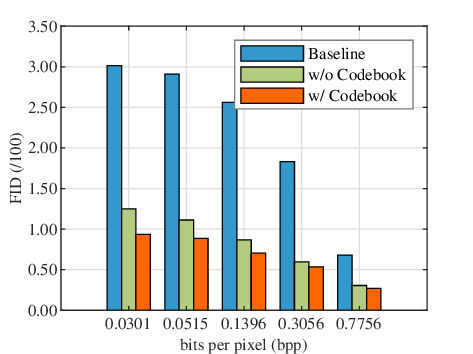}
		\end{minipage}	
	}%
	\subfigure 
	{
		\centering
		\begin{minipage}[t]{0.24\textwidth}
			\includegraphics[scale=0.6]{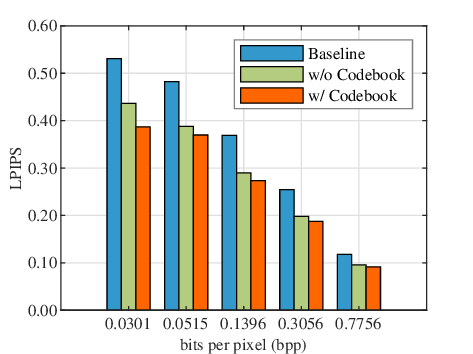}
		\end{minipage}	
	}%
	\centering
	\caption{Decoding performance of the baseline (ELIC) and the proposed Code-RSIC with and without codebook in a wide range of bitrates on the testing set of LoveDA.}
	\label{Fig: Hist_abla_codebook}
\end{figure}
\subsubsection{Effectiveness of Codebook}
To evaluate the effectiveness of the generated codebook, Fig. \ref{Fig: Hist_abla_codebook} shows the decoding performance of the baseline and the proposed Code-RSIC with and without the codebook in terms of FID and LPIPS under five values of \(\lambda \in \{4, 8, 32, 100, 450\} \times 10^{-4}\). It should be noted that here the option ``without codebook" refers to Stage III, where we only employ HPIN without fusing the high-quality prior of the decoder \(D_H\). From the results, it is evident that even without the codebook, the perception quality shows significant improvement compared to the baseline across a wide range of bitrates, thanks to the powerful capability of the developed Transformer-based hierarchical model and GAN technology. Particularly, after introducing the codebook, the perception quality achieves better results at all bitrates. \par

\begin{figure}[!htbp]
	\centering
	\includegraphics[width=0.44\textwidth]{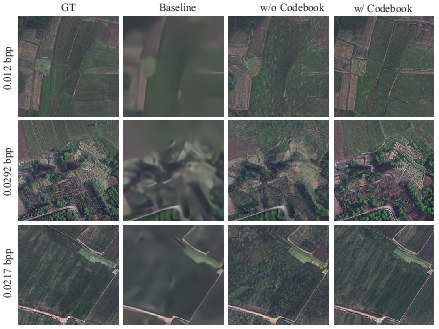}
	\caption{Decoding performance of the baseline and the proposed Code-RSIC with and without codebook at extremely low bitrates ($\lambda$=0.0004).}
	\label{Fig: Abaltion_codebook}
\end{figure}

To visualize the decoding results of each option, Fig. \ref{Fig: Abaltion_codebook} presents the decoded images of ``4241", ``4282", and ``4257" from the testing set of LoveDA at an extremely low bitrate (\textit{e.g.}, $\lambda=0.0004$). As shown in the results, the decoded images from the baseline suffer from severe blurring issues, while the proposed Code-RSIC without a codebook significantly improves the perceptual quality. However, it is clear that the proposed Code-RSIC presents more realistic decoding results after the introduction of the codebook. This improvement is due to the pre-trained codebook reducing the uncertainty in the ill-posed inverse problem of RS image decoding. \par
In brief, we can carefully demonstrate that the codebook plays an important role in decoding texture-rich RS images even at extremely low bitrates. In practice, it is only necessary to deploy the codebook on the decoding side of the RS image compression algorithm. This allows for the pre-storage of high-quality prior information, reducing the need for data stream transmission.

\subsection{Limitations}
In the experiments, comprehensive experimental results have demonstrated that the proposed Code-RSIC presents an impressive ability to decode texture-rich RS images, especially at extremely low bitrates. However, it remains challenging to achieve fidelity results, particularly at low bitrates. The failure cases shown in Fig. \ref{Fig: Abaltion_codebook} illustrate that, although the perceptual quality has been significantly improved, there are still noticeable fake texture details in the decoded images. This may be because there is minimal known information in the decoded images of the baseline, making it difficult for Code-RSIC to predict more accurate codes, resulting in less effective high-quality RS image decoding.

Moreover, the perceptual quality improvement of Code-RSIC is primarily because there are many inter-image similarity prior among various RS images, allowing the encoding of their high-level features to build a codebook. However, the performance is not outstanding for complex and variable image scenes. This problem may be alleviated by further extending the codebook space and expanding the training dataset.
\begin{table*}[!htbp]
	\renewcommand\arraystretch{1.2}
	\centering
	\caption{Comparison of the computational complexity of various learning-based image compression algorithms}
	\begin{tabular}{ccccccc}
		\toprule
		\multirow{2}{*}{Methods} &  \multirow{2}{*}{FLOPs (G)} & \multirow{2}{*}{Parameters (M)} & \multicolumn{2}{c}{CPU} & \multicolumn{2}{c}{GPU} \\
		\cmidrule(lr){4-5}   \cmidrule(lr){6-7}  &     &     & \multicolumn{1}{l}{Encoding Time (s)} & \multicolumn{1}{l}{Decoding Time (s)} & Encoding Time (s) & Decoding Time (s) \\
		\hline
		Entroformer (ICLR22) &  44.76 & 12.67 & 1.3482  & 0.5018  & 0.2762  & 0.0919  \\
		STF (CVPR22) & 99.83 & 33.35 &  1.1395  & 1.3175  & 0.1343  & 0.1879  \\
		ELIC (CVPR22) &  31.66 &  54.46 & 0.7329  & 0.8454  & 0.0402  & 0.0224  \\
		LIC-TCM (CVPR23) &  35.23 & 44.97 & 22.0959  & 21.0951  & 0.4637  & 0.4045  \\
		Code-RSIC (Our) &  133.99 & 126.07 & 0.7329  & 18.4962  & 0.0402  & 0.2270 		\\
		\bottomrule
	\end{tabular}%
	\label{tab:parameter_FLOP}%
\end{table*}%
\subsection{Model Complexity}
Assessing the computational complexity of compression algorithms is crucial for evaluating their practical feasibility and efficiency. Here, we conduct experiments on the workstation hardware, including an Intel Silver 4214R CPU with 2.40 GHz and a NVIDIA GeForce RTX 3090 Ti GPU, using the PyTorch 2.0.1 framework with CUDA 11.7. We employ the floating-point operations per second (FLOPS), model parameters, encoding time, and decoding time as our evaluation criteria. The experiment involved 100 images with a resolution of \(256 \times 256 \times 3\), and the average model complexity results are reported in Table \ref{tab:parameter_FLOP}. The results indicate that the proposed method exhibits high model complexity, which is primarily attributed to the global computations required by the Transformer blocks and the MCMs. \par

Furthermore, in terms of encoding and decoding time of the compression algorithms, LIC-TCM exhibits the slowest performance on the CPU, with encoding and decoding times of 22.0959s and 21.0951s, respectively. In contrast, the proposed Code-RSIC shows shorter encoding and decoding times of 0.7329s and 18.4962s, respectively. When leveraging the computational power of the GPU, the encoding and decoding times for LIC-TCM are reduced to 0.4637s and 0.4045s, while for Code-RSIC, they decrease to 0.0402s and 0.2270s.

In summary, it is evident that the proposed method has lower runtime complexity compared to LIC-TCM, despite having higher model FLOPs and parameters. This demonstrates the practical efficiency of our method in real-world applications. However, there are several avenues for future research. One potential direction is to optimize the Transformer block to reduce its computational burden further while maintaining high performance. Additionally, exploring more efficient codebook structures or alternative attention fuse mechanisms could help in achieving a better balance between model complexity and decoding performance. 

\section{Conclusion} \label{Sec:Conclusion}
In this paper, we propose a codebook-based remote sensing image compression (Code-RSIC) method by utilizing the inter-image similarity prior. Firstly, we employ VQGAN to learn a codebook with a large number of high-quality RS images. Then we introduce a Transformer-based prediction module to reduce the gap between the latent features of the decoded images from ELIC and the pre-trained codebook. Finally, we utilize Transformer blocks and multi-head cross-attention modules (MCMs) to develop a hierarchical prior integration network (HPIN), which leverages inter-image similarity prior in the built codebook for improved RS image quality. Extensive experimental results demonstrate the superiority of the proposed Code-RSIC in perceptual quality over existing state-of-the-art traditional and learning-based image compression algorithms. In the future, we will aim to extend the proposed method to enhance its generalization capabilities across various RS image scenarios, thereby further improving its practical value.

\bibliographystyle{IEEEtran}
\bibliography{reference.bib}


%

%



\ifCLASSOPTIONcaptionsoff
  \newpage
\fi

\end{document}